%% file: root.tex
\documentclass[letterpaper, 10 pt, conference]{ieeeconf}  

\IEEEoverridecommandlockouts                              

\input{packages}

\title{\LARGE \bf
Height Control and Optimal Torque Planning for Jumping With Wheeled-Bipedal Robots
}

\author{Yulun Zhuang$^{1}$, Yuan Xu$^{1}$, Binxin Huang$^{1}$, Mandan Chao$^{1}$, \\Guowei Shi$^{1}$, Xin Yang$^{1}$, Kuangen Zhang$^{2,3}$, Chenglong Fu$^{2}$
\thanks{This work was supported by the National Key R\&D Program of China [Grant 2018YFB1305400]; National Natural Science Foundation of China [Grant U1913205]; Guangdong Basic and Applied Basic Research Foundation [Grant 2020B1515120098]; Guangdong Innovative and Entrepreneurial Research Team Program [Grant 2016ZT06G587]; the Science, Technology and Innovation Commission of Shenzhen Municipality [Grant SGLH20180619172011638 and Grant ZDSYS20200811143601004]; and Centers for Mechanical Engineering Research and Education at MIT and SUSTech; Special Funds for the
Cultivation of Guangdong College Students’ Scientific and Technological Innovation[Grant pdjh2021c0041].}
\thanks{The first three authors contributed equally. The last two authors are co-corresponding authors.}
\thanks{
$^{1}$Yulun Zhuang, Yuan Xu, Binxin Huang, Mandan  Chao, Guowei Shi, Xin Yang are with Department of Mechanical and Energy Engineering, Southern University of Science and Technology, Shenzhen, 518055, China.(e-mail:z1079931505@gmail.com, xuyuan2966@gmail.com, asqrew@126.com)}
\thanks{$^{2}$Kuangen Zhang and Chenglong Fu are with Shenzhen Key Laboratory of Biomimetic Robotics and Intelligent Systems and Guangdong Provincial Key Laboratory of Human-Augmentation and Rehabilitation Robotics in Universities, Department of Mechanical and Energy Engineering, Southern University of Science and Technology, Shenzhen, 518055, China.(e-mail: kuangen.zhang@alumni.ubc.ca, fucl@sustech.edu.cn)}%
\thanks{$^{3}$Kuangen Zhang is also with the Department of Mechanical Engineering, The University of British Columbia, Vancouver, BC, V6T1Z4, Canada.}%
}

\begin{document}

\maketitle
\thispagestyle{empty}
\pagestyle{empty}

\begin{abstract}
This paper mainly studies the accurate height jumping control of wheeled-bipedal robots based on torque planning and energy consumption optimization. Due to the characteristics of underactuated, nonlinear estimation, and instantaneous impact in the jumping process, accurate control of the wheeled-bipedal robot's jumping height is complicated. In reality, robots often jump at excessive height to ensure safety, causing additional motor loss, greater ground 
reaction force and more energy consumption. To solve this problem, a novel wheeled-bipedal jumping dynamical model(W-JBD) is proposed to achieve accurate height control. It performs well but not suitable for the real robot because the torque has a striking step. Therefore, the Bayesian optimization for torque planning method(BOTP) is proposed, which can obtain the optimal torque planning without accurate dynamic model and within few iterations. BOTP method can reduce 82.3\% height error, 26.9\% energy cost with continuous torque curve. This result is validated in the Webots simulation platform. 
Based on the torque curve obtained in the W-JBD model to narrow the searching space, BOTP can quickly converge (40 times on average).
Cooperating W-JBD model and BOTP method, it is possible to achieve the height control of real robots with reasonable times of experiments.

\end{abstract}

\section{Introduction}

Over the past years, mobile robots have assumed a central role with the improvement of control methods and small-size actuator. Taking advantage of mobility\cite{8594484}, they offer the potential for applications on automatic patrolling, cleaning, warehousing, surveying and mapping. 

The traditional mobile robots including wheeled robots and bipedal robots\cite{LI2014188} are incompetent to deal with complex scenarios and challenging terrains such as gaps, stepping and stones. Combined with the high mobility efficiency\cite{8419761} from wheeled robots and the ability to deal with complicated terrains from bipedal robots\cite{7758092}, the wheeled-bipedal robots can complete flexible locomotion missions\cite{Shen2018ALC}. 

Previous research such as Handle and Ascento\cite{Klemm_2019}\cite{Klemm_2020}, realized balanced stance, dynamic height control and other fundamental control work in wheeled-bipedal robots, and developed hardware platforms that can be applied to complex motions. Ascento applied linear quadratic regulator (LQR) to the whole body control(WBC) in control. It is proposed to use the differential flatness of wheeled-bipedal robot model in the trajectories planning and LQR controller\cite{8724780} in tracking the planning curves\cite{chen2020underactuated}. However, the system applied with the LQR controller requires linearity, which significantly limits the model control. The robot model always contains non-linear terms due to unknown complexity, especially of wheeled-bipedal robot, which has underactuation\cite{4108935}\cite{407375} and coupling impeding the model control.

Although recent research has revealed important insights into the wheeled-bipedal robot's jumping, there currently exists no comprehensive theoretical method linking to optimize the jumping process\cite{5175424}\cite{6906613}. To this end, we establish a wheeled-bipedal jumping dynamical(W-JBD) model to optimize the height control. However, for practical applications, there are several factors hard to determine, such as the damping and friction of the joints \cite{dahl1976solid}. Hence, we deploy the Bayesian optimization for torque planning(BOTP) method based on a joint optimization framework to obtain the optimal torque planning to achieve accurate height control, protect motors and minimize energy cost. In conclusion, the proposed W-JBD model provides a solution for achieving accurate jumping control with wheeled-bipedal robots\cite{9339917}\cite{LI2020342}. Furthermore, cooperating with the BOTP method, the application of height control on the real robot is available.

\section{Models and Control}

The whole jumping process consists of five phases: squat, stance, take-off, flying, and landing. It should satisfy the requirement of more precisely jump height control, less energy consumption, and motor protection. The inverted pendulum is applied to the stance phase for keeping balance and being on speed in part A. Robot squat to wait for action, which is realized by dynamic height control in part B. There are two methods of jumping. One is the W-JBD model. Given a certain desired height, specific knee motors' rotational angles and torque $\tau(\theta)$ will be generated in part C. The other is the BOTP method in part D, which is more suitable for applying to real robots. The last phase is landing. With the intervening of the balance controller, the robot displays a hard landing and restores to stance pose. So that soft landing is the future work.

\begin{figure}[H]
	\centering
	{\includegraphics[width=0.9\linewidth]{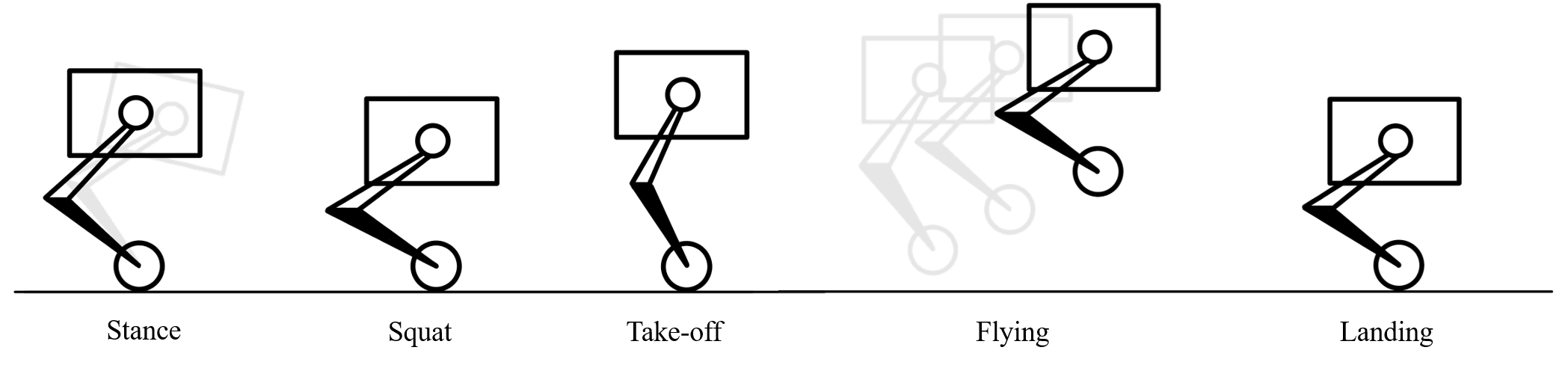}}
	\caption{\footnotesize Motion states}
	\label{fig:Motion state}
	\vspace{0px}
\end{figure}

\subsection{Stance Model}
In order to control the balance of the wheeled-bipedal robot, two knee joints’ angles maintain unchanged. A single inverted pendulum model abstracts the essential dynamics of wheeled-bipedal robot. Equations of the force and torque of wheels in steady state are given below.

{
\setlength\abovedisplayskip{1pt}
\setlength\belowdisplayskip{1pt}
\begin{equation}
    \left\{
        \begin{array}{lr}
        \vspace{1ex}
            F_x = f-m\Ddot{\alpha}r\\\vspace{1ex}
            \tau=\frac{1}{2}mr^2\Ddot{\alpha}+fr\\\vspace{1ex}
            2F_x=M\frac{d}{dt^2}(x+L\sin{\theta})\\\vspace{1ex}
            2F_y-Mg=M\frac{d}{dt^2}(L\cos{\theta})\\\vspace{1ex}
            ML^2\Ddot{\theta}=MgL\sin{\theta}-M\Ddot{\alpha}rL\cos{\theta}-2\tau
        \end{array}
    \right.
    \label{stance}
\end{equation}
}

We define state variables $\bm{x}=[\theta,\dot \theta]^T$ and control variable $\bm{u}=2\tau$. By solving equations \eqref{stance}, $\ddot{\theta}$ in $\dot{\bm x}$ is given as

\begin{equation}
    \ddot{\theta} =
    \frac{-\delta_1\delta_2\theta^2+\delta_1 Mgr + 6\delta_1 J_\omega g/r - 2\tau(\delta_2+Mr^2+6J_\omega)}
        {6J_\omega ML^2+\delta^2_1}
    \vspace{2ex}
\end{equation}
where $J_\omega=\frac{1}{2}mr^2$ is wheel's moment of inertia, and $\delta_1 = MLr\sin \theta$, $\delta_2 = MLr\cos \theta$.

When the robot is around the upright position where $\hat{\bm x}_\theta=[0,0]^T$, the model can be linearized by using Taylor expansion at that point.
The linearized model is given as
\begin{align}
    \hat{\bm A}
    &=\frac{\partial f}{\partial x_{\theta}}\Bigg|_{ \hat{x}_{\theta}}
    = 
    \begin{bmatrix}
        0 & 1 \\
        \frac{g(Mr^2+6J_\omega)}{6J_\omega L} & 0 \\
    \end{bmatrix}
    \\
    \hat{\bm B}
    &=\frac{\partial f}{\partial u}\Bigg|_{\hat{x}_{\theta}}
    = 
    \begin{bmatrix}
        0  \\
        -\frac{MLr+Mr^2+6J_\omega}{6J_\omega ML^2} \\
    \end{bmatrix}
\end{align}

We implement state feed back control to balance the robot by selecting the desired closed-loop eigenvalues and computing the gain Matrix $\bm K$ such that $eigs(\hat{\bm A}-\hat{\bm B}K) = s_{1,2}$.

\subsection{Squat Model}
The robot needs to squat before taking off, and after that, legs will crank to overcome obstacles. Therefore, dynamic height control is essential for the robot to complete the jumping task. The essene of dynamic height control is to plan the trajectory of the center of mass (CoM) of the robot\cite{liu2019dynamic}, and the CoM satisfies three specific constraints while squatting.
The constraints are
\begin{itemize}
    \item The robot base remains upright during squat
    \item The CoM of robot except wheels is always on the line of gravity of wheels
    \item The distance $h$ from the CoM of the robot base to the CoM of the wheels satisfies the kinematics constraints
\end{itemize}

\begin{figure}[H]
    \vspace{-8px}
	\centering
	{\includegraphics[width=0.35\linewidth]{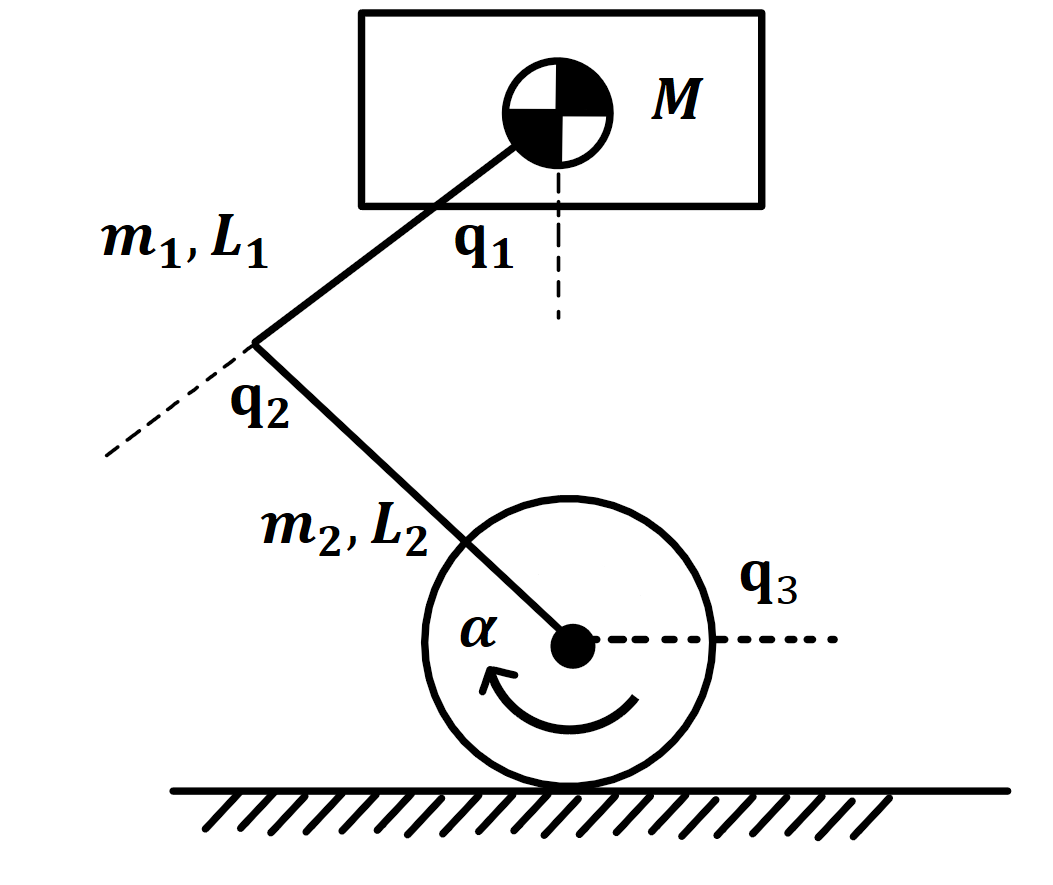}}
	\caption{\footnotesize 2D notations of the Kinematics model for wheeled-bipedal robot}
	\label{fig:Kinematics}
	\vspace{-12px}
\end{figure}

We establish a simplified kinematic model of the wheeled-bipedal robot shown in Fig.\ref{fig:Kinematics}

\begin{figure}[H]
    \vspace{-13px}
	\centering
	{\includegraphics[width=0.75\linewidth]{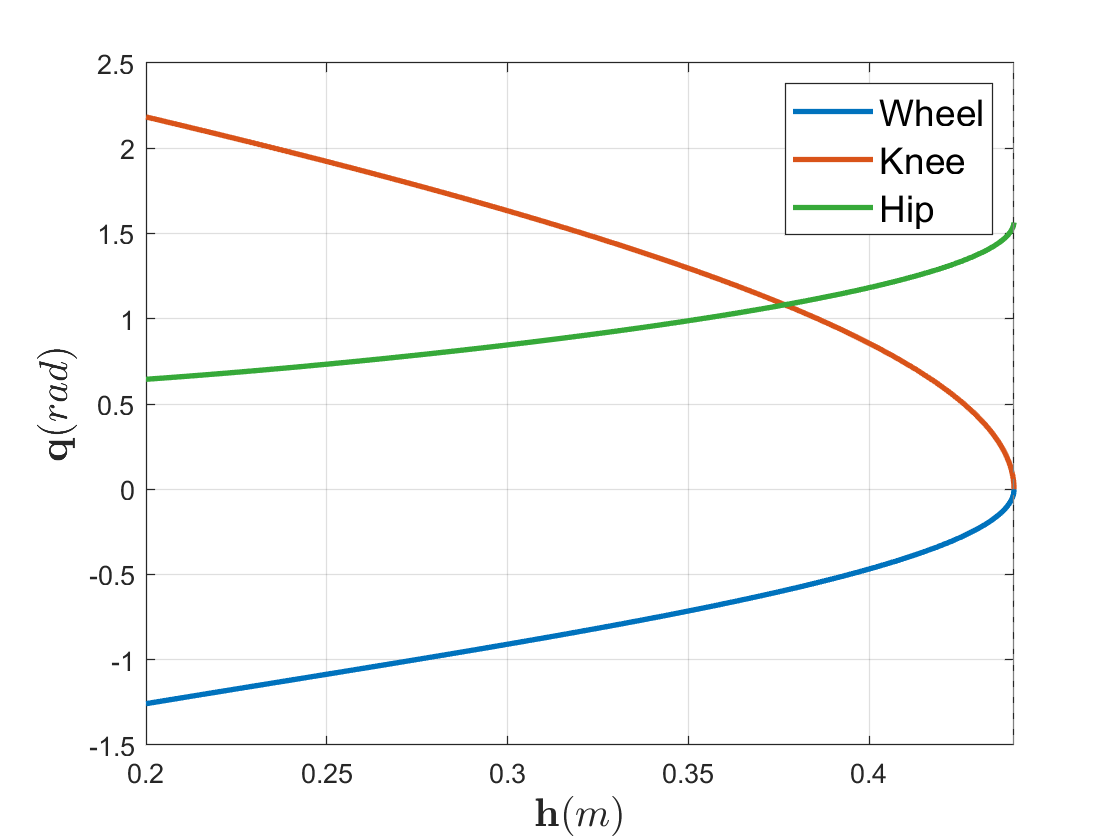}}
	\caption{\footnotesize The motion planning curve for squatting}
	\label{fig:squat}
	\vspace{-5px}
\end{figure}

We define the joint variables $\bm{q}=[q_1, q_2, q_3]^T$ be the angle of hip, knee and wheel motors respectively. We implement the squat controller by solving the inverse kinematics to obtain the control sequence of $\bm q$ in joint space and using position control of hip and knee motors to squat the robot. The relationship between $\bm q$ and h is shown in Fig.\ref{fig:squat}

\subsection{W-JBD Model}
To jump to a specified height, we model the robot's jumping process.
Some researches have used the SLIP model for robot jumping. However, it considers the robot a mass point at the top, ignoring other parts' mass.  During the flying phase, the robot will contract legs to overcome obstacles, which is unconsidered in the SLIP model. Therefore, this paper further proposes the W-JBD model, which has noteworthy improvement compared to SLIP model.
\begin{figure}[H]
	\centering
	{\includegraphics[width=0.6\linewidth]{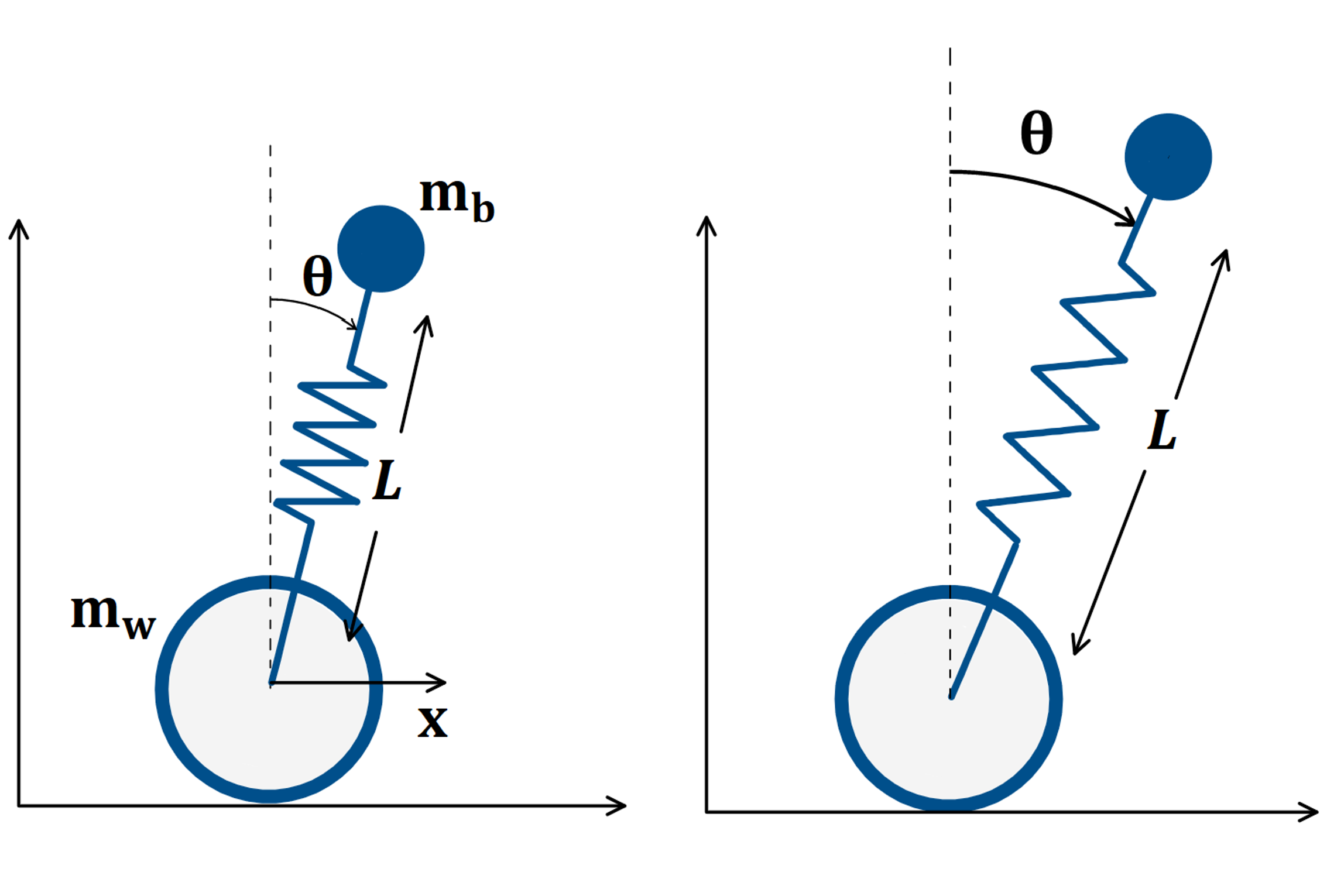}}
	\caption{\footnotesize The W-JBD Model of the Wheeled-Bipedal Robots}
	\label{fig:SLIP}
	\vspace{-13px}
\end{figure}
In the take-off phase, since the balance controller is and mature and stable, the robot body's pitch angle can be steered to a small value at the moment of jumping, and there is almost no speed in the pitch direction. i.e. $\theta\approx0$ and $\dot{\theta} \approx 0$ (Fig.\ref{fig:SLIP}). Therefore, we can further simplify the model only along Z-axis.
Firstly, we apply the SLIP model to the robot. Assuming the spring displacement $\Delta L$, the equivalent centroid mass $m_b$ (total mass except wheels'), the take-off velocity $\dot{L}$,  we can derive the equation with the energy conservation.
\begin{equation}
    \frac{1}{2}K_s\Delta L- m_b g\Delta L = \frac{1}{2}m_b\dot{L}^{2}
    \label{eqn: SLIP}
\end{equation}

\begin{figure}[H]
	\centering
	{\includegraphics[width=0.5\linewidth]{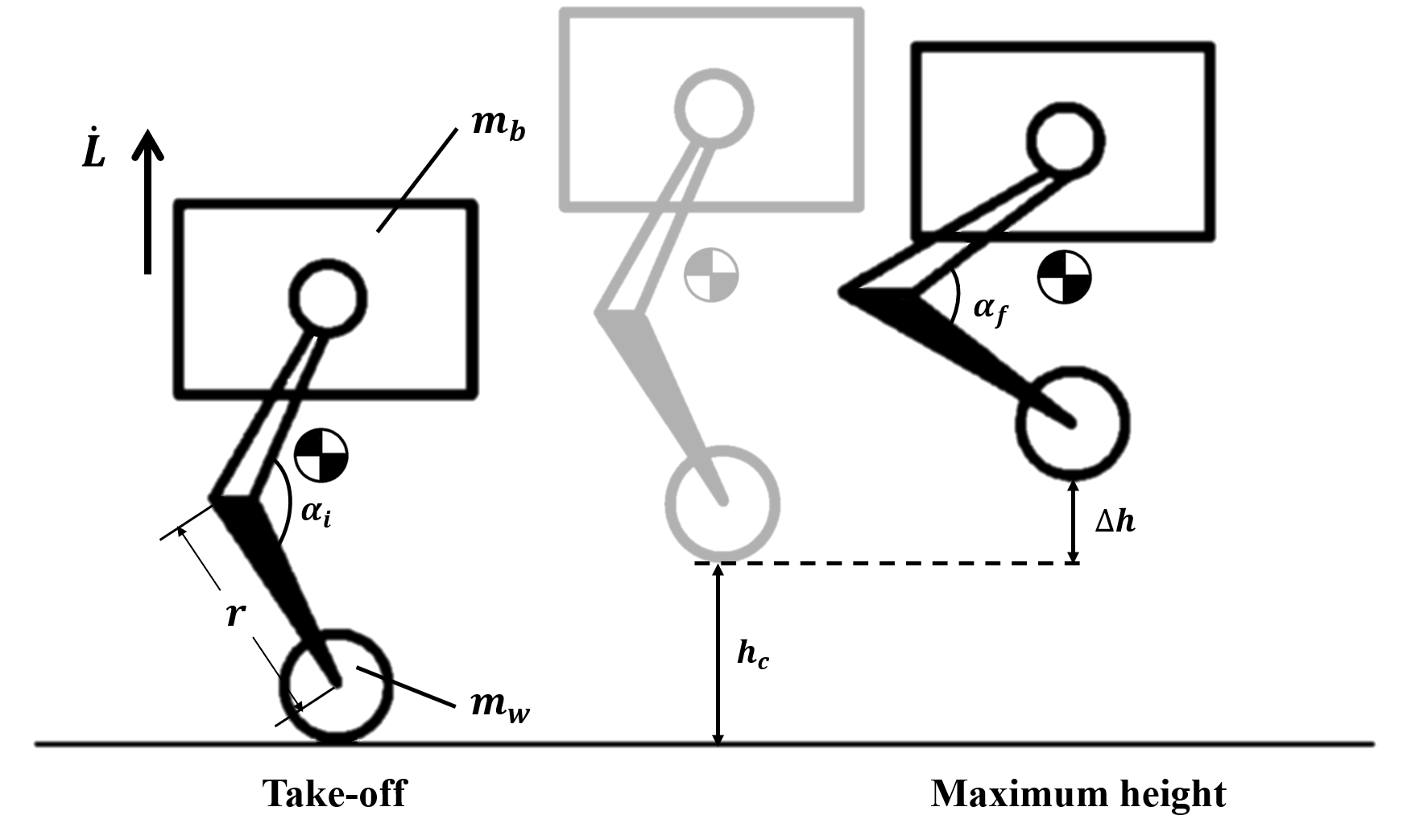}}
	\caption{\footnotesize The W-JBD Model of the Wheeled-Bipedal Robots}
	\label{fig:Takeoff}
	\vspace{-16px}
\end{figure}

The robot will leave the ground when the equivalent spring comes to the original length. The robot step into the flying phase after take-off, and the legs will crouch from angle $\alpha_i$ to angle $\alpha_f$ before $CoM$ reaching the highest point, which helps maximize the flying height to overcome the obstacle.
During the flying phase, the $CoM$ rising and the legs crouching are simultaneous but two independent processes. For analyzing more conveniently, two processes are drawn respectively. The parameters are shown in Fig.\ref{fig:Takeoff}. Finally, we can directly solve the height $h_w$, which is the wheels' distance off the ground.

\begin{equation}
    \left\{
        \begin{array}{lr}
              m_b \dot{L}^{2}=(m_\omega+m_b) V_{CoM} \vspace{1ex}\\
              V_{CoM}^{2} = 2g h_c \vspace{1ex}\\
              (m_\omega+m_b)\Delta h = 2m_br[ sin\frac{\alpha_i}{2}-sin\frac{\alpha_f}{2}] \vspace{1ex}\\
              h_w = h_c+\Delta h \vspace{1ex}
        \end{array}
    \right.
\end{equation}
finally, deriving the height of wheel
\begin{equation}
\vspace{1ex}
    h_\omega = \frac{1}{2g}(\frac{m_\omega+m_b}{m_b })^2+2r[sin{\frac{\alpha_i}{2}}-sin{\frac{\alpha_f}{2}}]\frac{m_b}{m_\omega+m_b}\vspace{1ex}
    \label{eqn: VBM}
\end{equation}

Substitute Eqn(\ref{eqn: VBM}) into Eqn(\ref{eqn: SLIP}), we can derive the relationship between designed spring displacement $\Delta \widetilde{L}$, spring stiffness $K_s$ and desired height $\widetilde{h}_w$
\begin{equation}
    K_s \Delta \widetilde{L}^{2}-2m_b g \Delta \widetilde{L} = 2g \widetilde{h}_w\frac{M^{2}}{m_b}-4\eta g r M
    \label{eqn: W-JBD}
\end{equation}
where $M=m_\omega+m_b$,
$\eta=sin\frac{\alpha_i}{2}-sin\frac{\alpha_f}{2}$,
$\Delta \widetilde{L}=L_0-2rsin\frac{\widetilde{\alpha}}{2}$ and $\widetilde{\alpha}$ is the corresponding designed angle of knee motors before jump. 

\subsection{BOTP Method}
In addition to the feedforward control based on the W-JBD model, this paper also proposes the BOTP method, which will plan motors' torque curve $\tau(t)$ according to the desired jumping height.
In Eqn(\ref{eqn: VBM}), reversely, if $h_w$ in the above formula is replaced by the desired height $\widetilde{h}_w$, the desired take-off velocity $\dot{L}_d$ can be obtained. The sensor can obtain the actual body speed of the model. When the vertical velocity reaches the desired take-off velocity $\dot{L}_d$ that we designed, the motors will stop applying torque, and then the robot will get off the ground. 

However, in the BOTP method, the motors' angle $\alpha_i$ varies from different torque curves $\tau(t)$, making the influence of cranking legs unpredictable, which will significantly affect the final jumping height. Therefore, the value of desired take-off velocity $\dot{L}_d$ should be different from the $\dot{L}$ calculated in W-JBD model. Since it is unpractical to derive the explicit relationship between $\tau(t)$ and the $\dot{L}_d$, we decide to use the Bayesian optimization to find the best solution, and more information is presented in the next part.

\section{Optimization and Simulation}

Model-based optimization can simultaneously combine the advantages of model and optimization, reduce model accuracy requirements, reduce the number of parameters required to be optimized, and decrease the optimization epochs.
In the W-JBD model, we simplify the robot's physical properties, but due to the estimation error of inertial matrix of legs, friction, and damping of motors, solving the optimal control sequence of input torque online is a complex constrained nonlinear programming problem (NLP). For this factor, we propose a simulation-based joint optimization framework, which combines the Webots simulator and the Advisor optimizer to achieve accurate control of the jumping process by simultaneously optimizing the control sequence of the input torque and the critical speed of the take-off phase. With this approach, we can obtain the optimal parameters of BOTP to control the robot's jumping height accurately and meet the actual constraints.

\begin{figure}[H]
	\centering
	{\includegraphics[width=0.7\linewidth]{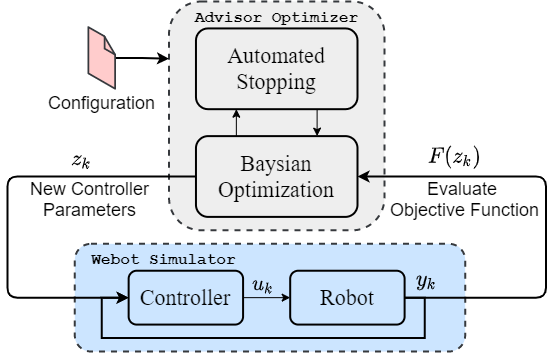}}
	\caption{\footnotesize The simulation-based joint optimization framework}
	\label{fig:trainframe}
	\vspace{-15px}
\end{figure}

\subsection{Bayesian Optimization}

Firstly, we build the optimization model. The optimization goal is the difference between the actual jumping height of the robot's wheels $h_w$ and the target height $\widetilde{h}_w$. The actual jumping height of the robot's wheels is calculated by the simulation physics engine. In the model and controller section, we abstract it into a nonlinear model. We hope to optimize the control sequence of input torque to minimize the jumping height error, that is, the most accurate jumping height. The constrained NLP problem is modeled as follows.

\begin{subequations}
    \vspace{-13px}
    \begin{align}
        \min \quad &\Vert h_w - \widetilde{h}_w \Vert^2\\
        s.t. \quad
            &\bm G(\bm q)\leq 0\\
            &\bm H(\bm q)\ddot{\bm q}+\bm C(\bm q,\dot{\bm q})=\bm B \bm u\\
            &\bm q\in \bm Q, \bm u\in \bm U
    \end{align}
    \label{eq:primal}
\end{subequations}
where $\bm G(\bm q)$ are constraints comes from dynamics, kinematics and power limits of the robot.

Furthermore, we use the square penalty function to convert the constrained NLP into an unconstrained NLP. While removing the constraints, the penalty function method punishes the infeasible solution of the constrained problem by modifying the objective function.
{
\setlength\abovedisplayskip{1pt}
\setlength\belowdisplayskip{1pt}
\begin{equation}
    \min \quad \| h_w - \widetilde{h}_w \|^2+\mu\sum_{i=1}^m p_i
    \label{eqn: dual}
\end{equation}
}
where $p_i=(\max\{0,g_i\})^2$ and $\mu$ is the penalty factor.

It is easier to experiment with the robot jumping task than to fully understand the jumping model. Manual adjustment of parameters through trial and error is time-consuming and inefficient. We can optimize the NLP problem as a black-box optimization problem. Black-box optimization is the task of optimizing an objective function $F(z_k)$ with a limited budget for evaluations. 

Bayesian optimization provides an elegant method of using information from previous search points to determine the next search point for solving the black-box optimization problem\cite{mockus1978application}, and has been shown its performance is better than other global optimization algorithms\cite{jones2001taxonomy}.

In this work, we use the well-implemented Bayesian Optimization Algorithm (BOA) of the Advisor optimizer to solve the optimization problem. Advisor is an open-source implementation of Google Vizier\cite{golovin2017google}, which is a hyperparameter tuning system for black-box optimization, especially used for auto machine learning (AutoML).

\subsection{Simulation Platform} 
\begin{figure}[h]
	\centering
	{\includegraphics[width=0.3\linewidth]{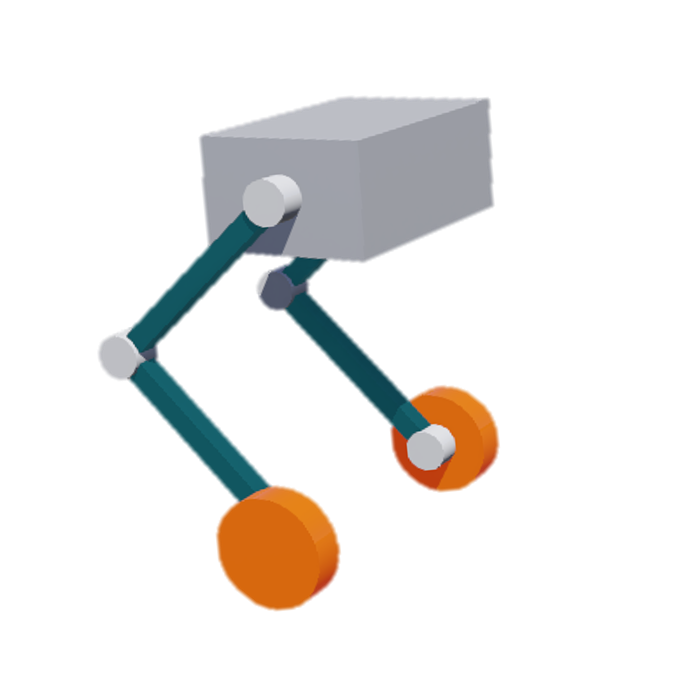}}
	\caption{\footnotesize Simplified model for simulation}
	\label{fig:Model&CAD}
	\vspace{-15px}
\end{figure}

The open-source Webots simulator based on a simplified wheeled-bipedal robot shown in Fig. \ref{fig:Model&CAD} was used to verify the overall model and controllers. The total mass of the robot is $7.8 kg$. The joint torque limit of the hip and knee motors is set to $35 N\cdot m$, and the joint torque limit of the wheel is set to $5 N\cdot m$, which accurately represents the actuator torque limit of the motor used in the actual robot. We added friction and damping coefficients to the joint motor in the model to make the simulation closer to actual conditions.

\subsection{Joint Optimization Framework}
The overall simulation-based optimization framework is set up by jointly combining Advisor optimizer and Webots simulator as shown in Fig.\ref{fig:trainframe}

Initially, we determine the optimizer setting (i.e., number of iterations, parameters specifications) in a configuration file, and then run the Advisor optimizer.

Advisor optimizer invokes BOA after a warm start (a certain number of random searches)\cite{jones2001taxonomy}, passes in new controller parameters for Webots simulator, and then evaluates the returned objective value to update the black-box model. Note that the automated stopping algorithm will decide whether to terminate the optimization early according to the Performance Curve Stopping Rule\cite{golovin2017google}.

In Webots simulator, the simulation engine receives the optimized controller parameters, and then calculates the optimal control sequence of input torque based on the proposed model. It quantitatively evaluates the robot's jumping performance (jumping height deviation and energy consumption under constraints) as the objective value of this optimization.

\section{Results}

\subsection{Jumping Performance}
\begin{figure}[H]
	\centering
	{\includegraphics[width=0.9\linewidth]{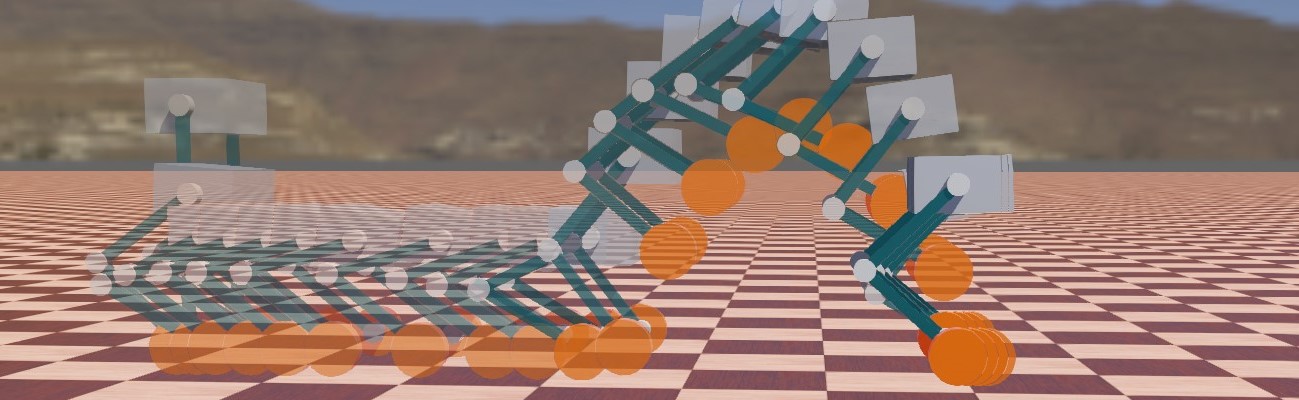}}
	\caption{\footnotesize The afterimage of jumping}
	\label{fig:canying}
	\vspace{-15px}
\end{figure}

\begin{figure}[H]
	\centering
	{\includegraphics[width=0.8\linewidth]{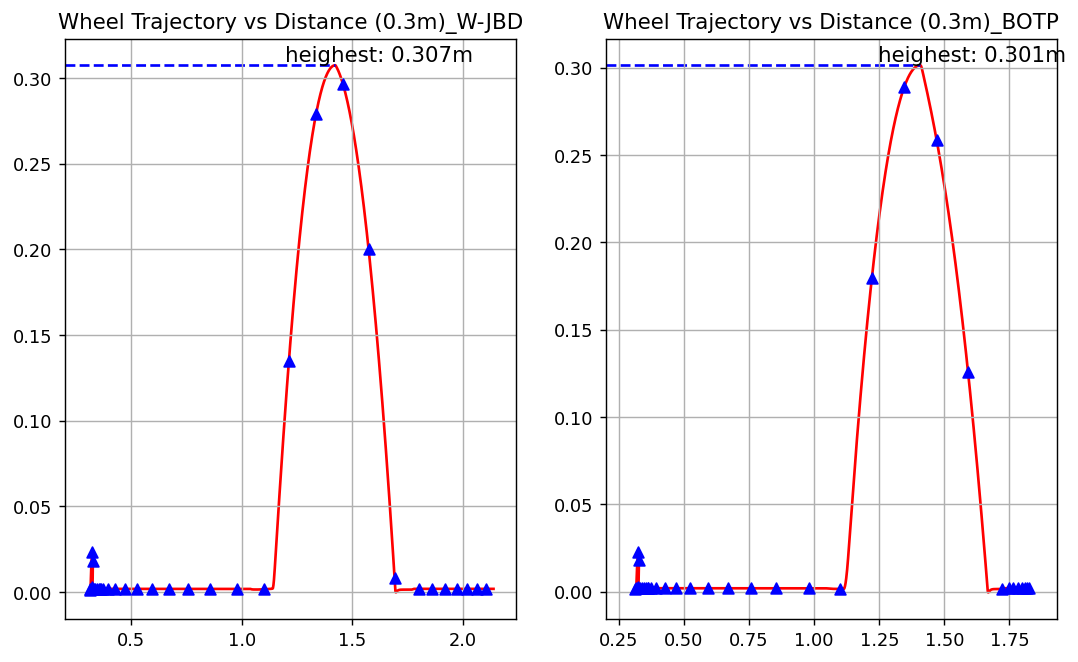}}
	\caption{\footnotesize The jumping trajectory of W-JBD and BOTP(desired height is 0.3m)}
	\label{fig:traj}
	\vspace{-5px}
\end{figure}

The afterimage shown in Fig.\ref{fig:canying} illustrates the whole process of jumping. The jumping performance is mainly measured by the deviation of jumping height. In comparation, both W-JBD and BOTP can reach a height deviation within 4\% (Fig.\ref{fig:traj}, the sampling rate is 10Hz).

\subsection{Optimization Performance}
\begin{figure}[H]
	\centering
	{\includegraphics[width=0.9\linewidth]{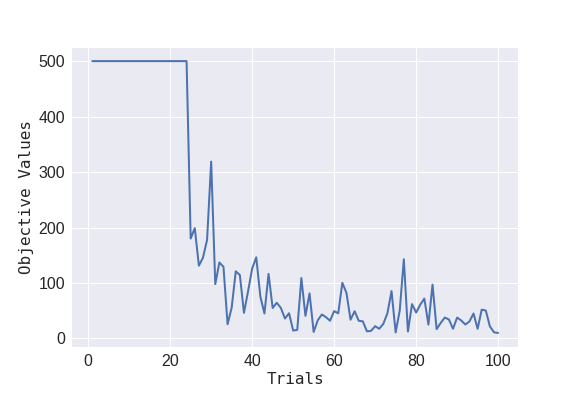}}
	\caption{\footnotesize Typical decreasing objective values during optimization}
	\label{fig:loss_h=2}
	\vspace{-10px}
\end{figure}

BOA can optimize the objective values of the given jumping task within 100 iterations and be close to convergence within 40 iterations (Fig.\ref{fig:loss_h=2}). By using the joint optimization framework, we optimize different jumping heights to obtain the control parameter sequence corresponding to the certain jumping height, and build the jumping parameter library of the robot to adapt to more general jumping tasks.

\begin{figure}[H]
	\centering
	\vspace{-5px}
	{\includegraphics[width=0.9\linewidth]{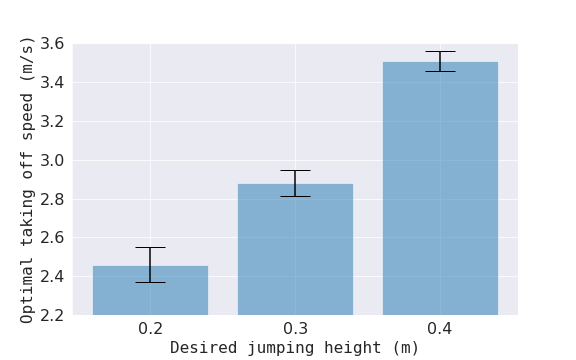}}
	\caption{\footnotesize Optimal taking off speed with different desired jumping height}
	\label{fig:opt_vel}
	\vspace{-5px}
\end{figure}

In the experiment, we analyze the optimization parameters of three different desired jumping heights ($h_w=0.2,0.3,0.4$) shown in Fig.\ref{fig:opt_vel}, the error bars represent the standard deviation of top ten sets of optimal parameters. This shows that the robot’s best optimal take-off speed has a clear positive correlation with desired jumping height.

\subsection{The Precision of Jumping Height}
\begin{figure}[h]
	\centering
    \vspace{-5px}
	{\includegraphics[width=0.8\linewidth]{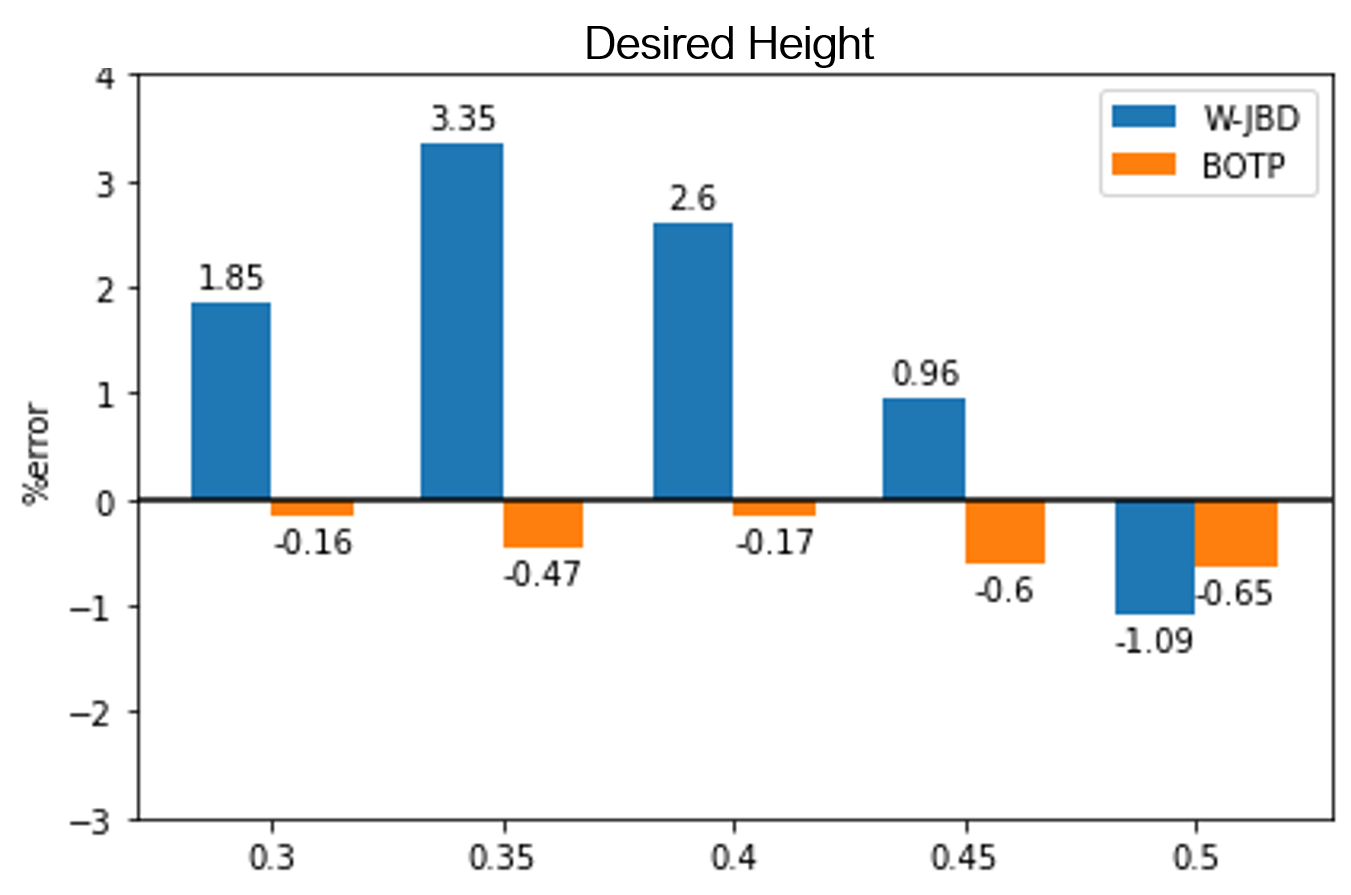}}
	\caption{\footnotesize The height error of W-JBD model and BOTP method with different desired height}
	\label{fig:height_err}
	\vspace{-5px}
\end{figure}

The first indicator of evaluating the performance is the accuracy of jumping height shown as Fig.\ref{fig:height_err}. It plots that the W-JBD model and BOTP method both show good performance in this indicator, as the error is limited within 4\% for W-JBD model(corresponding absolute error is limited at most 0.015m with maximum desired height 0.5m), and 0.2\% for BOTP method, which has better accuracy in jumping height.

\subsection{The Torque Trajectory}

\begin{figure}[H]
	\centering
	{\includegraphics[width=0.8\linewidth]{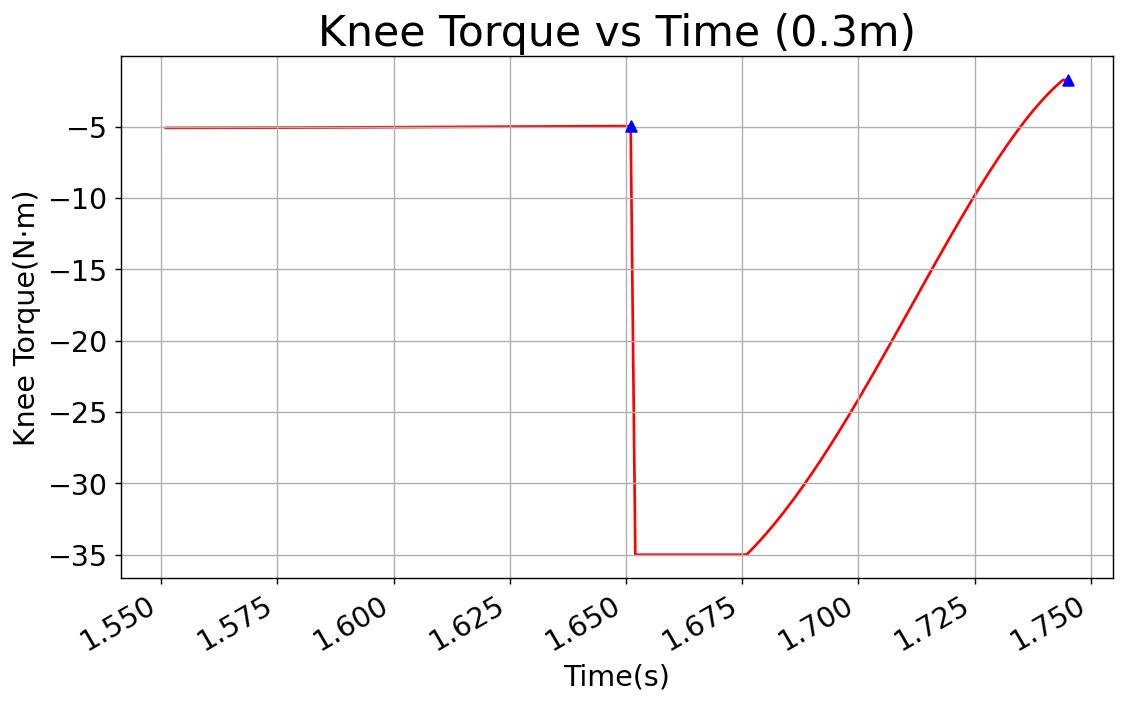}}
	\caption{\footnotesize The toque trajectory of W-JBD model in stance and take-off phase}
	\label{fig:slip_knee_torque}
	\vspace{-5px}
\end{figure}

Our experiments focus on the take-off phase optimization, i.e., the curve shown below between the blue points. Fig.\ref{fig:slip_knee_torque} illustrates the torque curve of the knee motors in the case of the W-JBD model with the desired height of 0.3m. Notice that there is a striking step from -4.96N·m to -35N·m at the first blue point, which may damage the motors when the torque curve is applied to a real robot.

\begin{figure}[H]
	\centering
	{\includegraphics[width=0.8\linewidth]{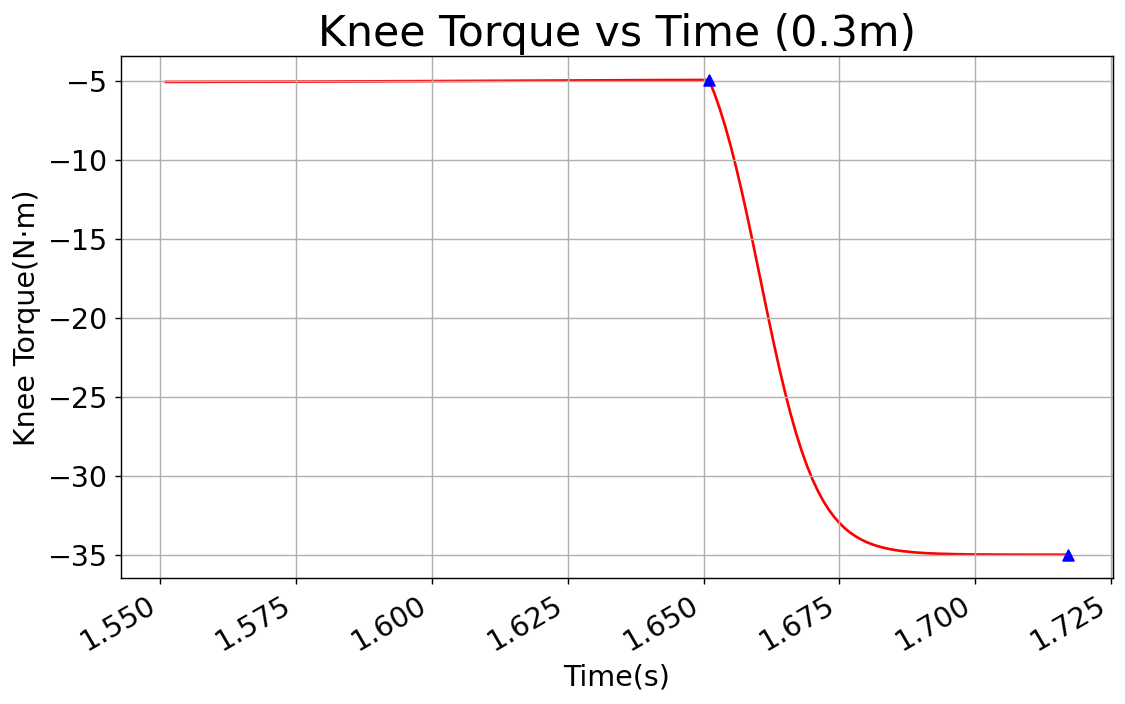}}
	\caption{\footnotesize The toque trajectory of BOTP method in stance and take-off phase}
	\label{fig:sigmoid_knee_torque}
	\vspace{-15px}
\end{figure}

BOTP is employed to find a proper torque curve $\tau(t)$, which is continuous and will not damage the motors. With W-JBD providing the torque curve that can determine a reasonable searching space for Bayesian optimization, significantly reducing the number of iterations of parameter optimization, the W-JBD model helps the BOTP method to work better, making the BOTP method workable on the real robot. We get a sample performance shown in Fig.\ref{fig:sigmoid_knee_torque} with desired height 0.3m.

\subsection{Energy Cost}
\begin{figure}[H]
	\centering
	{\includegraphics[width=0.8\linewidth]{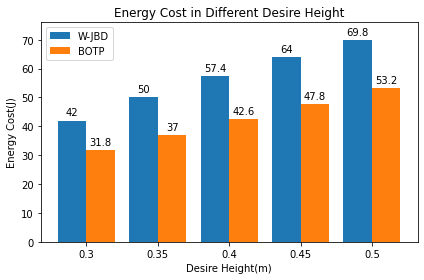}}
	\caption{\footnotesize The energy cost using W-JBD model and BOTP method with different desired height}
	\label{fig:energy}
	\vspace{-15px}
\end{figure}

\begin{figure}[H]
	\centering
	{\includegraphics[width=0.8\linewidth]{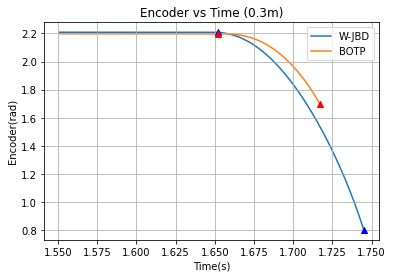}}
	\caption{\footnotesize The encoder value of knee motors via time, the curve between the points(red or blue respectfully) stands for the take-off phase}
	\label{fig:encoder}
	\vspace{-5px}
\end{figure}

Energy Cost is the work energy of the knee motors, which is transformed to gravitational potential energy and damping energy loss. Normally, the energy cost increases along with the height increment(Fig.\ref{fig:energy}). Analyzing the data in Fig.\ref{fig:energy}, BOTP takes less energy, though it needs the larger torque but its take-off phase traveled less stroke as shown in Fig.\ref{fig:encoder}.

\section{Conclusion}
In this paper, a novel wheeled-bipedal jumping dynamical (W-JBD) model is proposed for achieving accurate height control for jumping. The W-JBD model performs well on height control, but it does not consider the joint's torque limit so that the improper torque planning will harm motors, making it unavailable to apply to the real robot.  To optimize the torque planning,  a simulation-based joint optimization framework is developed. Cooperating with the W-JBD model, a Bayesian optimization for torque planning(BOTP) method is proposed and obtains the optimal torque planning with 40 epochs averagely, which is reasonable for applying on the real robot. The performance of the overall framework is verified with Webots simulation.

In the future, we aim to apply the BOTP method to optimize the landing phase of jumping, minimize the damage caused by ground impact. Furthermore, we will improve the BOTP method to reduce the number of iterations further.

\section{Acknowledgement}
The authors would like to thank Prof.Wei Zhang and Prof. Hua Chen for providing the Mini-NeZha model, and essential guidance in stance control of wheeled-bipedal robot.

\addtolength{\textheight}{-12cm}   



\newpage
\bibliography{ref}
\bibliographystyle{IEEEtran}
\end{document}

%% file: packages.tex
\let\labelindent\relax
\usepackage{times,multirow,float}
\usepackage{enumitem}
\usepackage{wrapfig}
\usepackage{graphicx}
\usepackage{epsfig,xspace,layout}
\usepackage{color}
\usepackage{amsfonts}
\usepackage{times}
\usepackage{amssymb}
\usepackage{hyperref}
\usepackage{amsmath,bm}
\usepackage{amsmath}
\usepackage{rotating}
\usepackage{graphicx}
\usepackage{epsfig}
\usepackage{mathrsfs}
\usepackage{mathtools}
\usepackage{makeidx}
\usepackage{multirow} 
\usepackage{dblfloatfix}
\usepackage{threeparttable}
\usepackage{dsfont}
\usepackage{cite}
\usepackage{xcolor}
\usepackage{lipsum}
\usepackage[font={small}]{caption}
\usepackage{algorithm}
\usepackage{algorithmic}
\usepackage{tikz}
\usetikzlibrary{shapes,arrows,positioning,calc}
\usepackage{siunitx}
\usepackage{multicol,lipsum}

\usepackage{url}
\usepackage{booktabs}
\usepackage{lineno}
\usepackage{hhline}
\usepackage[Symbol]{upgreek}

%% file: ref.bib
@article{Klemm_2019,
   title={Ascento: A Two-Wheeled Jumping Robot},
   ISBN={9781538660270},
   DOI={10.1109/icra.2019.8793792},
   journal={2019 International Conference on Robotics and Automation (ICRA)},
   publisher={IEEE},
   author={Klemm, Victor and Morra, Alessandro and Salzmann, Ciro and Tschopp, Florian and Bodie, Karen and Gulich, Lionel and Kung, Nicola and Mannhart, Dominik and Pfister, Corentin and Vierneisel, Marcus and et al.},
   year={2019},
   month={May}
}

@article{Klemm_2020,
   title={LQR-Assisted Whole-Body Control of a Wheeled Bipedal Robot With Kinematic Loops},
   volume={5},
   ISSN={2377-3774},
   DOI={10.1109/lra.2020.2979625},
   number={2},
   journal={IEEE Robotics and Automation Letters},
   publisher={Institute of Electrical and Electronics Engineers (IEEE)},
   author={Klemm, Victor and Morra, Alessandro and Gulich, Lionel and Mannhart, Dominik and Rohr, David and Kamel, Mina and de Viragh, Yvain and Siegwart, Roland},
   year={2020},
   month={Apr},
   pages={3745-3752}
}

@misc{chen2020underactuated,
      title={Underactuated Motion Planning and Control for Jumping with Wheeled-Bipedal Robots}, 
      author={Hua Chen and Bingheng Wang and Zejun Hong and Cong Shen and Patrick M. Wensing and Wei Zhang},
      year={2020},
      eprint={2012.06156},
      archivePrefix={arXiv},
      primaryClass={cs.RO}
}

@article{Shen2018ALC,
  title={A Lateral Control Method for Wheel-Footed Robot Based on Sliding Mode Control and Steering Prediction},
  author={Wei Shen and Zhichun Pan and M. Li and H. Peng},
  journal={IEEE Access},
  year={2018},
  volume={6},
  pages={58086-58095}
}

@inproceedings{golovin2017google,
  title={Google vizier: A service for black-box optimization},
  author={Golovin, Daniel and Solnik, Benjamin and Moitra, Subhodeep and Kochanski, Greg and Karro, John and Sculley, D},
  booktitle={Proceedings of the 23rd ACM SIGKDD International Conference on Knowledge Discovery and Data Mining},
  pages={1487--1495},
  year={2017}
}

@article{mockus1978application,
  title={The application of Bayesian methods for seeking the extremum},
  author={Mockus, Jonas and Tiesis, Vytautas and Zilinskas, Antanas},
  journal={Towards Global Optimization},
  volume={2},
  number={117-129},
  pages={2},
  year={1978}
}

@article{jones2001taxonomy,
  title={A taxonomy of global optimization methods based on response surfaces},
  author={Jones, Donald R},
  journal={Journal of Global Optimization},
  volume={21},
  number={4},
  pages={345--383},
  year={2001},
  publisher={Springer}
}

@inproceedings{liu2019dynamic,
  title={Dynamic height balance control for bipedal wheeled robot based on ROS-Gazebo},
  author={Liu, Tangyou and Zhang, Chao and Song, Shuang and Meng, Max Q-H},
  booktitle={2019 IEEE International Conference on Robotics and Biomimetics (ROBIO)},
  pages={1875--1880},
  year={2019},
  organization={IEEE}
}

@article{dahl1976solid,
  title={Solid friction damping of mechanical vibrations},
  author={Dahl, Philip R},
  journal={AIAA journal},
  volume={14},
  number={12},
  pages={1675--1682},
  year={1976}
}

@article{LI2014188,
title = {Control of a Quadruped Robot with Bionic Springy Legs in Trotting Gait},
journal = {Journal of Bionic Engineering},
volume = {11},
number = {2},
pages = {188-198},
year = {2014},
issn = {1672-6529},
doi = {https://doi.org/10.1016/S1672-6529(14)60043-3},
author = {Mantian Li and Zhenyu Jiang and Pengfei Wang and Lining Sun and Shuzhi {Sam Ge}},
}

@INPROCEEDINGS{7758092,
author={M. {Hutter} and C. {Gehring} and D. {Jud} and A. {Lauber} and C. D. {Bellicoso} and V. {Tsounis} and J. {Hwangbo} and K. {Bodie} and P. {Fankhauser} and M. {Bloesch} and R. {Diethelm} and S. {Bachmann} and A. {Melzer} and M. {Hoepflinger}},
booktitle={2016 IEEE/RSJ International Conference on Intelligent Robots and Systems (IROS)}, title={ANYmal - a highly mobile and dynamic quadrupedal robot},
year={2016},
volume={},
number={},
pages={38-44},
doi={10.1109/IROS.2016.7758092}}

@ARTICLE{8419761,
  author={Y. {Zhang} and L. {Zhang} and W. {Wang} and Y. {Li} and Q. {Zhang}},
  journal={IEEE Access}, 
  title={Design and Implementation of a Two-Wheel and Hopping Robot With a Linkage Mechanism}, 
  year={2018},
  volume={6},
  number={},
  pages={42422-42430},
  doi={10.1109/ACCESS.2018.2859840}}

@ARTICLE{5175424,
  author={I. {Poulakakis} and J. W. {Grizzle}},
  journal={IEEE Transactions on Automatic Control}, 
  title={The Spring Loaded Inverted Pendulum as the Hybrid Zero Dynamics of an Asymmetric Hopper}, 
  year={2009},
  volume={54},
  number={8},
  pages={1779-1793},
  doi={10.1109/TAC.2009.2024565}}

@INPROCEEDINGS{6906613,
  author={P. M. {Wensing} and D. E. {Orin}},
  booktitle={2014 IEEE International Conference on Robotics and Automation (ICRA)}, 
  title={Development of high-span running long jumps for humanoids}, 
  year={2014},
  volume={},
  number={},
  pages={222-227},
  doi={10.1109/ICRA.2014.6906613}}

@INPROCEEDINGS{8724780,
  author={A. {Ashraf} and W. {Mei} and L. {Gaoyuan} and M. M. {Kamal} and A. {Mutahir}},
  booktitle={2018 IEEE 4th International Conference on Control Science and Systems Engineering (ICCSSE)}, 
  title={Linear Feedback and LQR Controller Design for Aircraft Pitch Control}, 
  year={2018},
  volume={},
  number={},
  pages={276-278},
  doi={10.1109/CCSSE.2018.8724780}}

@INPROCEEDINGS{407375,
  author={M. W. {Spong}},
  booktitle={Proceedings of IEEE/RSJ International Conference on Intelligent Robots and Systems (IROS'94)}, 
  title={Partial feedback linearization of underactuated mechanical systems}, 
  year={1994},
  volume={1},
  number={},
  pages={314-321 vol.1},
  doi={10.1109/IROS.1994.407375}}

@INPROCEEDINGS{4108935,
  author={M. {Park} and D. {Chwa} and S. {Hong}},
  booktitle={2006 SICE-ICASE International Joint Conference}, 
  title={Decoupling Control of A Class of Underactuated Mechanical Systems Based on Sliding Mode Control}, 
  year={2006},
  volume={},
  number={},
  pages={806-810},
  doi={10.1109/SICE.2006.315338}}

@INPROCEEDINGS{8594484,
  author={X. {Li} and H. {Zhou} and H. {Feng} and S. {Zhang} and Y. {Fu}},
  booktitle={2018 IEEE/RSJ International Conference on Intelligent Robots and Systems (IROS)}, 
  title={Design and Experiments of a Novel Hydraulic Wheel-Legged Robot (WLR)}, 
  year={2018},
  volume={},
  number={},
  pages={3292-3297},
  doi={10.1109/IROS.2018.8594484}}

@ARTICLE{9339917,  author={Li, Jiehao and Wang, Junzheng and Peng, Hui and Hu, Yingbai and Su, Hang},  journal={IEEE Transactions on Systems, Man, and Cybernetics: Systems},   title={Fuzzy-Torque Approximation-Enhanced Sliding Mode Control for Lateral Stability of Mobile Robot},   year={2021},  volume={},  number={},  pages={1-10},  doi={10.1109/TSMC.2021.3050616}}

@article{LI2020342,
    title = {Neural fuzzy approximation enhanced autonomous tracking control of the wheel-legged robot under uncertain physical interaction},
    journal = {Neurocomputing},
    volume = {410},
    pages = {342-353},
    year = {2020},
    issn = {0925-2312},
    doi = {https://doi.org/10.1016/j.neucom.2020.05.091},
    author = {Jiehao Li and Junzheng Wang and Hui Peng and Longbin Zhang and Yingbai Hu and Hang Su}
    }
